\documentclass{article} 
\usepackage{iclr2027_conference,times}

\usepackage{amsmath,amsfonts,bm}

\def\eqref#1{equation~\ref{#1}}

\def\1{\bm{1}}

\DeclareMathAlphabet{\mathsfit}{\encodingdefault}{\sfdefault}{m}{sl}
\SetMathAlphabet{\mathsfit}{bold}{\encodingdefault}{\sfdefault}{bx}{n}

\usepackage{hyperref}
\usepackage{url}
\usepackage{amssymb}
\usepackage{booktabs}
\usepackage{graphicx}
\usepackage{array}
\title{\raggedright
PhyRestore: Physics-Structured \\
Latent-Factor Restoration}

\author{
Ahmed Shafee \\
Department of Computer Science\\
Adams State University\\
Alamosa, CO, USA\\
\texttt{aashafee@adams.edu}
\And
Chayan Lahiri \\
Department of Geosciences\\
Adams State University\\
Alamosa, CO, USA\\
\texttt{chayanlahiri@adams.edu}
}

\iclrfinalcopy 
\begin{document}

\maketitle

\begin{abstract}
Estimating temporal soil-loss change is challenging when physically meaningful input factors are noisy or corrupted, particularly because substantial changes are rare relative to the large number of locations exhibiting little change. We study this problem through the Revised Universal Soil Loss Equation (RUSLE) and introduce PhyRestore, a physics-structured latent-factor restoration framework. Rather than directly predicting soil-loss change or correcting a degraded physical estimate, PhyRestore restores corrupted physical factors and reconstructs temporal change through the known physical relationship.  
We evaluate PhyRestore in a watershed-scale bitemporal raster setting under isolated and simultaneous corruption of rainfall erosivity and cover management, comparing it with the degraded RUSLE estimate and Direct RF, XGBoost, MLP, and CNN models. Factor restoration improves high-magnitude recovery when the corrupted factors remain identifiable, but its advantage weakens under joint corruption, sparse positive extremes, and factor values outside the training support.

\end{abstract}

\section{Introduction}
\label{sec:introduction}

Because direct soil-erosion measurements are typically limited to localized plots or field sites and are difficult to extrapolate reliably, watershed-scale assessments commonly combine empirical models with geospatial observations to produce spatially distributed estimates \citep{stroosnijder2005measurement,boixfayos2006measuring,phinzi2019assessment,benavidez2018review}. The Revised Universal Soil Loss Equation (RUSLE) is widely used for this purpose, integrating factors representing rainfall erosivity, soil erodibility, topography, cover management, and support practices \citep{renard1997rusle,benavidez2018review}. Although its factorized formulation is interpretable and practical when dense field observations are unavailable, its estimates remain sensitive to errors in the underlying factors \citep{benavidez2018review,phinzi2019assessment}.

We study this problem in a temporal setting using co-registered factor rasters from 2017 and 2022 over a Colorado watershed. The prediction target is the signed difference between the clean RUSLE estimates for the two years and therefore represents a RUSLE-derived temporal soil-loss proxy rather than independently measured soil-loss change. When a time-varying factor is corrupted at either date, the resulting error propagates through the corresponding annual estimate and can distort both the magnitude and direction of the estimated temporal change.

Three learning pathways are possible: directly predict the clean temporal target, learn a residual or gated correction to the degraded RUSLE estimate, or restore the corrupted factor before applying the known RUSLE relationship. These pathways differ in whether learning replaces the analytical estimate, corrects it after error propagation, or operates on the corrupted factor itself.

The temporal soil-loss changes are strongly imbalanced: most locations exhibit changes near zero, while large positive or negative changes are comparatively rare. Learning the final temporal target is therefore dominated by small-change samples, making rare high-magnitude changes particularly difficult to recover \citep{ribeiro2020imbalanced,yang2021delving} and allowing strong aggregate performance over the full spatial support to obscure substantial errors in the changes of greatest magnitude.


These considerations motivate PhyRestore, which estimates the clean value of a corrupted RUSLE factor from bitemporal spatial context and reconstructs signed temporal change through RUSLE. We evaluate it under isolated and simultaneous factor corruption against the degraded analytical estimate and Direct, formula-correction, and trust-correction strategies implemented with RF, XGBoost, MLP, and CNN models.


Our contributions are:
\begin{itemize}
    \item We formulate signed temporal soil-loss prediction under controlled corruption of year-specific physical factors and compare analytical, Direct, formula-correction, trust-correction, and factor-restoration pathways under a common spatial evaluation protocol.
    \item We introduce \textbf{PhyRestore}, a physics-structured framework that restores corrupted physical factors from bitemporal spatial observations and reconstructs temporal change through the known RUSLE relationship.
    \item We characterize when factor-level restoration improves rare high-magnitude recovery and when its advantage weakens because corrupted factors are insufficiently recoverable or lie outside the training support.
\end{itemize}

\section{Related Work}
\label{sec:related_work}

\textbf{RUSLE-Based Soil-Loss Estimation and Temporal Analysis.} RUSLE uses factors representing rainfall erosivity, soil erodibility, topography, cover management, and support practices to estimate spatial soil-loss patterns \citep{renard1997rusle,benavidez2018review}, supporting regional and continental assessments and analyses of environmental change \citep{panagos2015assessment,borrelli2017global}. Remote sensing and geographic information systems increase factor availability \citep{li2023soil}, while machine learning has been used to estimate erosion susceptibility, analyze driving factors, and improve RUSLE input layers \citep{ge2023soil,samarinas2024soil}. Because these applications remain sensitive to factor quality \citep{benavidez2018review}, we examine signed temporal-change prediction when specific year-dependent factors are corrupted.

\textbf{Physics-Guided and Hybrid Scientific Machine Learning.} Scientific machine learning incorporates domain knowledge through theory-guided modeling \citep{karpatne2017theory}, differential-equation constraints \citep{raissi2019physics}, and hybrid analytical--learned systems \citep{karniadakis2021physics,willard2022integrating}. PhyRestore follows the hybrid paradigm but uses RUSLE as a fixed reconstruction rather than an optimization constraint, restoring corrupted physical factors before analytical reconstruction rather than learning a post-hoc correction after factor error has propagated through RUSLE.

\textbf{Learning with Corrupted or Missing Covariates.} Prior work addresses imperfect inputs through robust prediction, denoising, and imputation, including marginalized corrupted-feature methods that account for stochastic covariate perturbations during training \citep{vanDerMaaten2013corrupted} and GAIN, which reconstructs missing values from partially observed data \citep{yoon2018gain}. 
Unlike generic imputation, PhyRestore supervises factor-specific restoration and restores each affected factor separately under simultaneous corruption.

\textbf{Imbalanced Regression and Rare Extreme Recovery.}
Imbalanced-regression objectives and aggregate metrics can be dominated by densely sampled target regions, motivating weighting, resampling, specialized objectives, and representation learning for rare values \citep{ribeiro2020imbalanced,yang2021delving}. 
Here, temporal changes concentrate near zero; PhyRestore addresses this imbalance indirectly by restoring the corrupted factor rather than reweighting the final target.

\section{Temporal Prediction under Factor Corruption}
\label{sec:problem}

\subsection{Temporal RUSLE-Derived Target}
\label{sec:temporal_target}

Let $\mathbf{s}\in\Omega$ denote a valid spatial location and $t\in\{2017,2022\}$ the observation year. RUSLE represents soil loss using rainfall erosivity $R$, soil erodibility $K$, the topographic factor $LS$, cover management $C$, and the support-practice factor $P$. In our study, $R$ and $C$ are year specific, $K$ and $LS$ are shared across both years, and $P=1$ \citep{Ahmet_Mediterranean}. 
The factor rasters are derived from PRISM precipitation for $R$~\citep{daly2002prism}, USDA SSURGO soils for $K$~\citep{usda_ssurgo}, USGS 3DEP elevation data for $LS$~\citep{usgs_3dep}, and Sentinel-2 land-cover data for $C$~\citep{sentinelData}.
The annual RUSLE-derived soil-loss proxy is therefore

\begin{equation}
    A_t(\mathbf{s})
    =
    R_t(\mathbf{s})K(\mathbf{s})LS(\mathbf{s})C_t(\mathbf{s}).
    \label{eq:annual_rusle}
\end{equation}

The prediction target is the signed change between the clean 2022 and 2017 estimates:

\begin{equation}
    T(\mathbf{s})
    \equiv
    \Delta A_{\mathrm{clean}}(\mathbf{s})
    =
    A_{2022}(\mathbf{s})-A_{2017}(\mathbf{s})
    =
    K(\mathbf{s})LS(\mathbf{s})
    \bigl[
        R_{2022}(\mathbf{s})C_{2022}(\mathbf{s})
        -
        R_{2017}(\mathbf{s})C_{2017}(\mathbf{s})
    \bigr].
    \label{eq:clean_delta}
\end{equation}

Positive values indicate an increase in the RUSLE-derived soil-loss proxy from 2017 to 2022, whereas negative values indicate a decrease. Thus, $T(\mathbf{s})$ is a factor-derived reference quantity rather than an independently observed field measurement of soil loss.

\subsection{Controlled Factor Corruption}
\label{sec:factor_corruption}



We distinguish the clean factors defining the fixed target from the observations available to the predictor. We corrupt $R_{2022}$ and $C_{2017}$ because they are distinct year-specific factors appearing in opposite terms of Eq.~\ref{eq:clean_delta}; their combination therefore perturbs both terms. Uncertainty in the static $K$ and $LS$ factors is outside this study's scope.

Rainfall erosivity is degraded using multiplicative Gaussian noise:

\begin{equation}
    \widetilde{R}_{2022}(\mathbf{s})
    =
    R_{2022}(\mathbf{s})
    \bigl(1+\epsilon_R(\mathbf{s})\bigr),
    \qquad
    \epsilon_R(\mathbf{s})
    \sim
    \mathcal{N}(0,0.15^2).
    \label{eq:r_corruption}
\end{equation}

The 2017 cover-management factor is degraded using additive Gaussian noise followed by clipping to its valid range:

\begin{equation}
    \widetilde{C}_{2017}(\mathbf{s})
    =
    \operatorname{clip}
    \bigl(
        C_{2017}(\mathbf{s})+\epsilon_C(\mathbf{s}),
        0,
        1
    \bigr),
    \qquad
    \epsilon_C(\mathbf{s})
    \sim
    \mathcal{N}(0,0.08^2).
    \label{eq:c_corruption}
\end{equation}

Perturbations are generated independently across raster cells before spatial samples are extracted. A fixed realization is reused across spatial folds, model seeds, model families, and prediction strategies, ensuring that all methods receive identical degraded observations. We consider four conditions: \textit{Clean}, with no corruption; \textit{$R_{2022}$}, with only 2022 rainfall erosivity corrupted; \textit{$C_{2017}$}, with only 2017 cover management corrupted; and \textit{Mixed}, with both factors corrupted. All untargeted factors retain their clean values. Corruption affects only the predictor inputs; the reference target in Eq.~\ref{eq:clean_delta} remains fixed across conditions.

\subsection{Prediction under Corrupted Factors}
\label{sec:corrupted_prediction}

Let $\widetilde{\mathbf{X}}(\mathbf{s})$ denote the bitemporal factor observations available around location $\mathbf{s}$ under a given condition. The task is to estimate the fixed clean target $T(\mathbf{s})$ without access to the clean value of any corrupted factor. The following pathways differ in how learning is combined with the known RUSLE relationship.

\textbf{Degraded analytical prediction.} The analytical reference evaluates RUSLE directly using the available factors:

\begin{equation}
    F(\mathbf{s})
    \equiv
    \Delta \widetilde{A}_{\mathrm{formula}}(\mathbf{s})
    =
    \widetilde{A}_{2022}(\mathbf{s})
    -
    \widetilde{A}_{2017}(\mathbf{s}),
    \label{eq:degraded_formula}
\end{equation}

where each $\widetilde{A}_{t}(\mathbf{s})$ is computed from the factor values available at year $t$. Under corruption, factor error therefore propagates directly through RUSLE.

\textbf{Learned temporal prediction.} Direct prediction bypasses the analytical estimate, Formula Correction adds a learned residual to it, and Trust Correction modulates that residual using a learned bounded gate:

\begin{align}
    \widehat{T}_{\mathrm{Direct}}(\mathbf{s})
    &=
    f_{\theta}
    \bigl(
        \widetilde{\mathbf{X}}(\mathbf{s})
    \bigr),
    \label{eq:direct_prediction}
    \\
    \widehat{T}_{\mathrm{Correction}}(\mathbf{s})
    &=
    F(\mathbf{s})
    +
    \delta_{\phi}
    \bigl(
        \widetilde{\mathbf{X}}(\mathbf{s})
    \bigr),
    \label{eq:formula_correction}
    \\
    \widehat{T}_{\mathrm{Trust}}(\mathbf{s})
    &=
    F(\mathbf{s})
    +
    \alpha_{\phi}
    \bigl(
        \widetilde{\mathbf{X}}(\mathbf{s})
    \bigr)
    \delta_{\phi}
    \bigl(
        \widetilde{\mathbf{X}}(\mathbf{s})
    \bigr),
    \qquad
    0
    \leq
    \alpha_{\phi}
    \bigl(
        \widetilde{\mathbf{X}}(\mathbf{s})
    \bigr)
    \leq
    1.
    \label{eq:trust_correction}
\end{align}

Thus, Direct prediction learns the final clean target without using $F(\mathbf{s})$, whereas Formula Correction and Trust Correction operate after factor error has propagated through RUSLE. In Trust Correction, the gate controls how strongly the learned residual modifies the degraded analytical prediction.

\textbf{Factor-level restoration.} Because the controlled degradation affects identifiable physical factors, an alternative is to restore the corrupted factor and apply the known RUSLE relationship afterward. For a corrupted factor $q$, this pathway is

\begin{equation}
    \widetilde{\mathbf{X}}(\mathbf{s})
    \xrightarrow{\;\text{factor restoration}\;}
    \widehat{q}(\mathbf{s})
    \xrightarrow{\;\text{RUSLE reconstruction}\;}
    \widehat{T}(\mathbf{s}).
    \label{eq:factor_restoration_pathway}
\end{equation}

\section{PhyRestore: Physics-Structured Latent-Factor Restoration}
\label{sec:phyrestore}

\begin{figure*}[t]
    \centering
    \includegraphics[width=\textwidth]{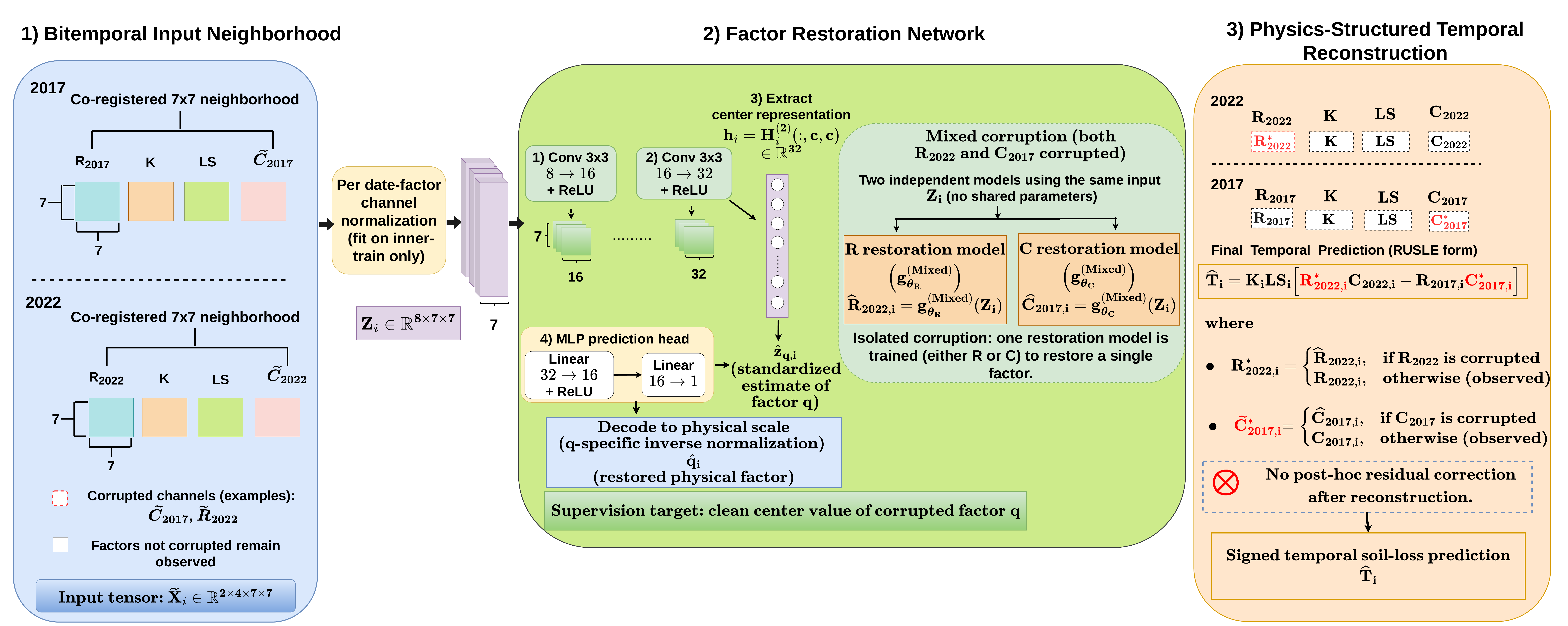}
    \caption{PhyRestore factor restoration and RUSLE-based temporal reconstruction. Under Mixed corruption, separate models restore $R_{2022}$ and $C_{2017}$.}
    \label{fig:phyrestore_overview}
\end{figure*}

PhyRestore learns in physical-factor space, restoring corrupted factors before deterministic reconstruction of the temporal target through RUSLE. Figure~\ref{fig:phyrestore_overview} summarizes the pipeline.

\subsection{Bitemporal Representation and Factor Restoration}
\label{sec:factor_restoration}

For each valid center location $i$, we extract co-registered $7\times7$ neighborhoods from 2017 and 2022. Each date contains the four RUSLE channels $[R,K,LS,C]$, producing the condition-specific bitemporal observation

\begin{equation}
    \widetilde{\mathbf{X}}_i
    \in
    \mathbb{R}^{2\times4\times7\times7}.
    \label{eq:bitemporal_input}
\end{equation}

A corrupted factor appears only in its degraded form; its clean value is not supplied through another input pathway. Separate normalization statistics are fitted for each date--factor channel using only the inner-training partition and held fixed during inner validation and outer-test inference. The dates are then concatenated along the channel dimension:

\begin{equation}
    \widetilde{\mathbf{X}}_i
    \longrightarrow
    \mathbf{Z}_i
    \in
    \mathbb{R}^{8\times7\times7}.
    \label{eq:reshape_input}
\end{equation}

Let $\mathcal{Q}_d$ denote the set of corrupted factors under condition $d$:

\begin{equation}
    \mathcal{Q}_d
    =
    \begin{cases}
        \varnothing, & d=\mathrm{Clean},\\
        \{R_{2022}\}, & d=R_{2022},\\
        \{C_{2017}\}, & d=C_{2017},\\
        \{R_{2022},C_{2017}\}, & d=\mathrm{Mixed}.
    \end{cases}
    \label{eq:corrupted_factor_set}
\end{equation}

For each $q\in\mathcal{Q}_d$, PhyRestore learns a factor-specific function that predicts the clean center value of $q$:

\begin{equation}
    \widehat{q}_i
    =
    g_{\theta_q}^{(d)}(\mathbf{Z}_i).
    \label{eq:factor_restore}
\end{equation}

Under Mixed corruption, separate restorers for $R_{2022}$ and $C_{2017}$ are trained using the same jointly corrupted observations:

\begin{equation}
    \widehat{R}_{2022,i}
    =
    g_{\theta_R}^{(\mathrm{Mixed})}(\mathbf{Z}_i),
    \qquad
    \widehat{C}_{2017,i}
    =
    g_{\theta_C}^{(\mathrm{Mixed})}(\mathbf{Z}_i).
    \label{eq:mixed_restorers}
\end{equation}

Restorers trained under isolated corruption are not reused under Mixed corruption because the available input information differs between these conditions.

\subsection{Restoration Network and Factor-Level Supervision}
\label{sec:restoration_network}

All factor restorers use the same 6,353-parameter convolutional architecture. Two unit-padded $3\times3$ convolutions preserve the $7\times7$ spatial dimensions and map the eight input channels to 16 and then 32 feature maps. The center representation is passed through a $32\rightarrow16\rightarrow1$ prediction head:

\begin{align}
    \mathbf{H}^{(1)}_i
    &=
    \operatorname{ReLU}
    \left(
        \operatorname{Conv}^{8\rightarrow16}_{3\times3}(\mathbf{Z}_i)
    \right),
    \qquad
    \mathbf{H}^{(2)}_i
    =
    \operatorname{ReLU}
    \left(
        \operatorname{Conv}^{16\rightarrow32}_{3\times3}
        (\mathbf{H}^{(1)}_i)
    \right),
    \label{eq:restore_cnn}
    \\
    \mathbf{h}_i
    &=
    \mathbf{H}^{(2)}_i(:,c,c)
    \in
    \mathbb{R}^{32},
    \label{eq:center_representation}
    \\
    \widehat{z}_{q,i}
    &=
    \mathbf{W}_2
    \operatorname{ReLU}
    \left(
        \mathbf{W}_1\mathbf{h}_i+\mathbf{b}_1
    \right)
    +\mathbf{b}_2,
    \label{eq:factor_head}
\end{align}

where $c$ is the center index of the spatial window. Using the same architecture for every restored factor prevents differences in restoration behavior from arising from factor-specific model capacity.

Factor targets are standardized using the mean $\mu_q$ and standard deviation $\sigma_q$ of the clean center values in the corresponding inner-training partition. The network is trained using Huber loss with $\delta=1$, and its predictions are decoded to physical units before reconstruction:

\begin{align}
    z_{q,i}
    &=
    \frac{q_i-\mu_q}{\sigma_q},
    \label{eq:factor_standardization}
    \\
    \mathcal{L}_q
    &=
    \frac{1}{N_{\mathrm{train}}}
    \sum_{i=1}^{N_{\mathrm{train}}}
    \ell_{\mathrm{Huber}}
    \left(
        \widehat{z}_{q,i},
        z_{q,i}
    \right),
    \qquad
    \delta=1,
    \label{eq:factor_loss}
    \\
    \widehat{q}_i
    &=
    \mu_q+\sigma_q\widehat{z}_{q,i}.
    \label{eq:factor_decode}
\end{align}

Checkpoint selection uses factor-level mean absolute error in the original physical units on the inner-validation partition. Neither the temporal target nor the held-out outer-test region is used for checkpoint selection.

\subsection{Physics-Structured Temporal Reconstruction}
\label{sec:temporal_reconstruction}

No additional learned transformation is applied after restoration. For each condition, reconstruction uses the restored value when a factor is corrupted and the observed value otherwise:

\begin{equation}
    R^{*}_{2022,i}
    =
    \begin{cases}
        \widehat{R}_{2022,i}, & R_{2022}\in\mathcal{Q}_d,\\
        R_{2022,i}, & R_{2022}\notin\mathcal{Q}_d,
    \end{cases}
    \qquad
    C^{*}_{2017,i}
    =
    \begin{cases}
        \widehat{C}_{2017,i}, & C_{2017}\in\mathcal{Q}_d,\\
        C_{2017,i}, & C_{2017}\notin\mathcal{Q}_d.
    \end{cases}
    \label{eq:reconstruction_factors}
\end{equation}

The final prediction follows directly from the temporal RUSLE relationship:

\begin{equation}
    \widehat{T}^{\mathrm{PhyRestore}}_i
    =
    K_iLS_i
    \bigl[
        R^{*}_{2022,i}C_{2022,i}
        -
        R_{2017,i}C^{*}_{2017,i}
    \bigr].
    \label{eq:phyrestore_prediction}
\end{equation}

Equation~\ref{eq:phyrestore_prediction} covers isolated $R_{2022}$ corruption, isolated $C_{2017}$ corruption, and Mixed corruption. Under Clean observations, no restoration model is required, and the prediction reduces to the clean temporal RUSLE difference. No learned residual or post-hoc correction follows reconstruction; the final estimate is determined entirely by the restored factors and the known RUSLE relationship.

\section{Experiments}
\label{sec:experiments}

\subsection{Experimental Setup}
\label{sec:experimental_setup}

\textbf{Data, baselines, and conditions.} We evaluate temporal soil-loss prediction using co-registered 2017 and 2022 RUSLE-factor rasters over a Colorado watershed, with a common valid support of 14,225 locations and $7\times7$ spatial neighborhoods. The clean signed target $T$ and the Clean, isolated $R_{2022}$, isolated $C_{2017}$, and Mixed corruption conditions are defined in Sections~\ref{sec:temporal_target} and~\ref{sec:factor_corruption}. We compare the degraded analytical estimate with the Direct, Formula Correction, and Trust Correction strategies from Section~\ref{sec:corrupted_prediction}, using MLP, random-forest (RF), XGBoost, and CNN models. The complete comparison covers three spatial outer folds and five seeds (101--105).

\textbf{Targeted high-magnitude controls.}
We additionally test seven controls that retain $T$ as the prediction target while modifying the spatial representation, loss, or output structure. RF7 replaces the baseline CNN's $3\times3$--$3\times3$ convolutions, whose theoretical receptive field is $5\times5$, with $5\times5$--$3\times3$ convolutions spanning the full $7\times7$ neighborhood. MLP37 augments 19 center-pixel inputs of the baseline MLP (MLP19)---eight bitemporal factor values, eight validity indicators, and three availability/corruption indicators---with 18 neighborhood summaries: the mean, standard deviation, and valid fraction of six factor surfaces over the 48 surrounding cells. 

MAG5-LINEAR-Q95 retains the MLP19 architecture but applies sign-symmetric magnitude weighting. On each inner-training split,
\[
s=q_{95,\mathrm{train}}(|T|),\qquad
w_i^{\mathrm{raw}}
=1+4\operatorname{clip}(|T_i|/s,0,1),\qquad
w_i=\frac{w_i^{\mathrm{raw}}}
{\operatorname{mean}_{\mathrm{train}}(w^{\mathrm{raw}})}.
\]
Thus, raw weights increase linearly from 1 to 5 and saturate at the inner-training 95th percentile of $|T|$. RF7, MLP37, and MAG5 are confirmatory controls evaluated over all three outer folds, five seeds, and four conditions, yielding complete out-of-fold predictions for each control.

The remaining controls are development-only experiments evaluated on folds 0 and 2 with seed 101 under all four conditions, using inner-validation data only. RankSim regularizes the Direct MLP representation using rank similarity \citep{gong2022ranksim}, with $\gamma=3\times10^{-4}$ and $\lambda=2$. Auxiliary classification adapts joint classification--regression learning \citep{pintea2023step}: signed-$T$ quartiles define four inner-training classes, and a shared MLP trunk feeds regression and classification heads trained with
$\mathcal{L}_{\mathrm{Huber}}+\beta\mathcal{L}_{\mathrm{CE}}$, where $\beta=0.0657$; only the regression output is used at inference.

SER-MAD adapts segmentation-based expert regression \citep{rashiwa2026segmentation}. Three MLP experts model lower, central, and upper target regimes separated at
$m\pm1.5\sigma_{\mathrm{MAD}}$, where
$m=\operatorname{median}(T)$ and
$\sigma_{\mathrm{MAD}}=1.4826\operatorname{median}(|T-m|)$ on the inner-training split. Deployable routing uses an ordinary Direct MLP prediction, whereas oracle routing uses the true $T$ to test whether perfect regime information would make specialization effective. Empty regimes fall back to the central expert.

The bitemporal Transformer applies self-attention \citep{vaswani2017attention} to 49 spatial tokens, each containing the four factor values from both dates. It uses 32-dimensional embeddings, two encoder layers, four attention heads, and the contextualized center token for regression.

\textbf{Protocol and metrics.}
For each seed and condition in the primary and confirmatory experiments, predictions from the three held-out outer regions are concatenated into an out-of-fold prediction over all 14,225 locations. Fitting, normalization, early stopping, and checkpoint selection use only the corresponding training data, and matched methods receive the same frozen corruption realization. Neural models use Huber loss with $\delta=1$, validation-based early stopping, and Adam; MAG5 instead selects checkpoints using its weighted validation objective.

We report MAE over the full support and the fixed TAIL90 and TAIL95 subsets. TAIL90 contains the 1,423 locations with
$|T|\geq0.0012514$, while TAIL95 contains the 712 locations with
$|T|\geq0.0019710$. Because TAIL95 contains 61 positive and 651 negative changes, we also evaluate TAIL95+ and TAIL95$-$. Comparisons with the zero predictor $\widehat{T}=0$, direction accuracy, and the prediction-to-target standard-deviation ratio distinguish low aggregate error from recovery of temporal-change sign and magnitude.

\subsection{Direct Prediction versus Formula-Anchored Correction}
\label{sec:strategy_results}

Table~\ref{tab:strategy_full} evaluates whether retaining the degraded analytical estimate as a prediction anchor improves robustness.

\begin{table}[t]
\centering
\caption{Full-support MAE ($\times10^{-3}$), reported as mean $\pm$ standard deviation across five seeds or once for seed-invariant methods. Bold marks the best learned strategy within each family.}
\label{tab:strategy_full}
\small
\setlength{\tabcolsep}{4.5pt}
\begin{tabular}{lccc}
\toprule
Method & $R_{2022}$ & $C_{2017}$ & Mixed \\
\midrule
Degraded Formula
& $10.271$
& $318.866$
& $318.865$ \\
\midrule
MLP--Direct
& $\mathbf{2.007 \pm 0.623}$
& $\mathbf{1.717 \pm 0.856}$
& $\mathbf{1.932 \pm 0.656}$ \\
MLP--Correction
& $6.076 \pm 0.884$
& $176.305 \pm 21.667$
& $173.110 \pm 26.649$ \\
MLP--Trust
& $6.045 \pm 0.809$
& $145.850 \pm 11.079$
& $150.191 \pm 12.054$ \\
\midrule
RF--Direct
& $\mathbf{0.981 \pm 0.005}$
& $\mathbf{1.048 \pm 0.006}$
& $\mathbf{0.979 \pm 0.011}$ \\
RF--Correction
& $7.127 \pm 0.085$
& $212.359 \pm 9.410$
& $210.358 \pm 9.581$ \\
RF--Trust
& $7.087 \pm 0.067$
& $186.493 \pm 6.035$
& $188.427 \pm 6.068$ \\
\midrule
XGB--Direct
& $\mathbf{0.913}$
& $\mathbf{0.905}$
& $\mathbf{0.927}$ \\
XGB--Correction
& $4.764$
& $117.242$
& $117.663$ \\
XGB--Trust
& $4.850$
& $112.904$
& $114.652$ \\
\midrule
CNN--Direct
& $\mathbf{1.094 \pm 0.265}$
& $\mathbf{1.163 \pm 0.165}$
& $\mathbf{1.278 \pm 0.335}$ \\
CNN--Correction
& $10.472 \pm 0.132$
& $245.604 \pm 25.101$
& $361.743 \pm 27.592$ \\
CNN--Trust
& $10.342 \pm 0.080$
& $283.415 \pm 21.067$
& $338.152 \pm 23.496$ \\
\bottomrule
\end{tabular}
\end{table}

Direct prediction has lower full-support MAE than Formula Correction and Trust Correction in all 12 family--corruption combinations. For stochastic learners, the ordering holds in all five seeds of every comparison and also holds seed-wise on TAIL95. Formula-anchored strategies nevertheless repair part of the analytical error in some cases: under $C_{2017}$ corruption, XGBoost reduces MAE from $0.318866$ to $0.117242$ with Formula Correction and $0.112904$ with Trust Correction. However, Trust is not consistently better than Formula Correction, and CNN correction can approach or exceed the degraded analytical error. Retaining the corrupted formula as an anchor therefore remains substantially less robust than predicting $T$ directly.

\subsection{Robustness Does Not Imply High-Magnitude Recovery}
\label{sec:direct_tail_results}

Because $T$ is concentrated near zero, all four primary Direct families have higher full-support MAE than the zero predictor under corrupted observations. Their tail behavior differs: MLP retains the strongest signed-tail signal, whereas RF, XGBoost, and CNN compress their predictions more strongly toward zero. Under $C_{2017}$ corruption on TAIL95, their prediction-to-target standard-deviation ratios are $0.301$, $0.054$, $0.014$, and $0.020$, respectively, while direction accuracies are $0.796$, $0.417$, $0.451$, and $0.022$. Thus, low aggregate error does not establish recovery of high-magnitude sign or variation. Table~\ref{tab:tail_controls} summarizes the targeted controls and their evaluation scope.

\begin{table}[t]
\centering
\caption{Controls targeting high-magnitude recovery while retaining $T$ as the prediction target. Confirmatory controls use three folds and five seeds; development controls use folds 0 and 2 and seed 101 with inner-validation data only.}
\label{tab:tail_controls}
\small
\setlength{\tabcolsep}{4pt}
\renewcommand{\arraystretch}{1.08}
\begin{tabular}{
>{\raggedright\arraybackslash}p{2.5cm}
>{\raggedright\arraybackslash}p{5.1cm}
>{\raggedright\arraybackslash}p{5.2cm}}
\toprule
Control & Intervention & Main outcome \\
\midrule
RF7
& Full theoretical $7\times7$ CNN receptive field
& No material signed-tail recovery relative to Direct CNN. \\
\addlinespace[2pt]
MLP37
& Eighteen summaries of the surrounding 48 cells
& No reproducible improvement over center-only MLP19. \\
\addlinespace[2pt]
MAG5
& Sign-symmetric magnitude-weighted regression
& No reproducible tail gain; full-support MAE worsens in all conditions. \\
\midrule
RankSim
& Rank-aware representation regularization
& Representation ordering improves, but TAIL90 and TAIL95 errors consistently worsen. \\
\addlinespace[2pt]
Auxiliary classification
& Four-class signed-target auxiliary task
& Does not establish a tail-regression gain. \\
\addlinespace[2pt]
SER-MAD
& Lower/central/upper regression experts
& Neither deployable nor oracle routing improves the evaluated TAIL95 subsets. \\
\addlinespace[2pt]
Bitemporal Transformer
& Attention over 49 paired-date spatial tokens
& No consistent positive-tail advantage after convergence. \\
\bottomrule
\end{tabular}
\end{table}

Among the confirmatory controls, RF7 yields mean TAIL95 MAEs of $0.005624$, $0.005591$, $0.005591$, and $0.005588$ under Clean, $R_{2022}$, $C_{2017}$, and Mixed observations, respectively, without materially improving signed-tail recovery over Direct CNN. MLP37 likewise provides no reproducible advantage over MLP19, with corresponding TAIL95 MAEs of $0.010718$, $0.012408$, $0.004714$, and $0.006525$. MAG5 worsens mean full-support MAE in every condition, and its isolated tail improvements do not meet the prespecified reproducibility criterion of improvement in at least four of five seeds.


In development, RankSim improves mean Spearman agreement by $0.188$ but worsens every evaluated TAIL90 and TAIL95 result. Auxiliary classification achieves approximately $0.20$ macro-F1 without improving tail regression, while neither deployable nor oracle SER-MAD routing improves TAIL95.

The Transformer's initial 40-epoch study was insufficient for convergence: seven of eight runs reached the maximum epoch. With a 200-epoch maximum, all eight runs converged and selected checkpoints between epochs 8 and 74. The converged Transformer improves over Direct MLP on Fold~2 TAIL95$-$ under $C_{2017}$ corruption ($0.004616$ versus $0.006761$), but is worse under $R_{2022}$ ($0.004826$ versus $0.004012$) and Mixed corruption ($0.004756$ versus $0.004172$), and improves none of the four Fold~0 TAIL95+ comparisons. Additional bitemporal attention therefore does not consistently resolve the tail failure.


These results motivate changing the learned target rather than only the loss or representation. PhyRestore restores the corrupted physical factors and then reconstructs temporal change using the known RUSLE relationship.

\subsection{Factor Restoration Improves High-Magnitude Recovery}
\label{sec:phyrestore_results}

\begin{figure}[t]
    \centering
    \includegraphics[width=\textwidth]{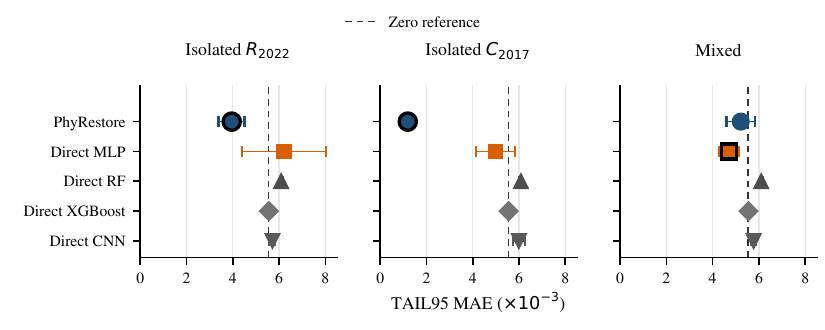}
    \caption{TAIL95 MAE ($\times10^{-3}$). Error bars show standard deviation across five seeds; dashed lines mark the zero predictor. The degraded formula is omitted because of its substantially larger scale.}
    \label{fig:tail95_main}
\end{figure}

A preliminary evaluation found recoverable factor-level signal, motivating full PhyRestore evaluation over three folds and five seeds. Because Formula Correction and Trust Correction are dominated on TAIL95 by their matched Direct models in every family and corrupted condition, Figure~\ref{fig:tail95_main} compares PhyRestore with the stronger Direct baselines and zero predictor.

Under isolated $R_{2022}$ corruption, PhyRestore obtains $0.003954$ TAIL95 MAE versus $0.005561$ for the strongest Direct baseline, a 28.9\% reduction. Under isolated $C_{2017}$ corruption, it obtains $0.001193$ versus $0.004989$, a 76.1\% reduction. Relative to the degraded formula, the reductions are 81.8\% and 99.8\%, respectively. PhyRestore outperforms Direct MLP in four of five runs for $R_{2022}$ and all five for $C_{2017}$, and outperforms RF, XGBoost, and CNN in all five runs under both isolated conditions.

Under Mixed corruption, Direct MLP is strongest, obtaining $0.004707$ versus $0.005217$ for PhyRestore. PhyRestore remains better on average than the degraded formula, zero predictor, RF, XGBoost, and CNN, but outperforms Direct MLP in only one of five seeds. TAIL90 exhibits the same pattern: PhyRestore is strongest under isolated corruption, while Direct MLP is strongest under Mixed corruption.

\textbf{Full-support accuracy.}
Tail improvements do not always extend to the full distribution. Under $R_{2022}$ corruption, PhyRestore obtains $0.001159\pm0.000064$ full-support MAE, compared with $0.000913$ for XGBoost, $0.000981$ for RF, and $0.000706$ for zero. Under Mixed corruption, the corresponding values are $0.001207\pm0.000063$, $0.000927$, $0.000979$, and $0.000706$. Under $C_{2017}$ corruption, however, PhyRestore is also strongest on full support, reaching $0.000065\pm0.000006$. The near-zero majority can therefore favor compressed predictions globally even when they recover high-magnitude changes poorly.

\subsection{Signed Extremes and Limits of Factor Recoverability}
\label{sec:recoverability_results}

Under isolated corruption, PhyRestore improves both signed TAIL95 subsets. For $R_{2022}$, its TAIL95+ and TAIL95$-$ MAEs are $0.005473$ and $0.003811$, versus the strongest Direct errors of $0.009031$ and $0.005191$. For $C_{2017}$, it obtains $0.006710$ and $0.000676$, versus $0.009238$ and $0.004542$. Under Mixed corruption, however, PhyRestore obtains $0.011995$ on TAIL95+, compared with $0.009031$ for Direct RF and $0.009487$ for zero, and $0.004582$ on TAIL95$-$, slightly above Direct MLP's $0.004189$. Moreover, under isolated $C_{2017}$ corruption, its positive-tail prediction-to-target standard-deviation ratio remains only $0.061$ despite the MAE improvement.

\textbf{Factor restoration depends on inner-training support.} Pooling outer-test predictions across seeds and relevant conditions, $R_{2022}$ restoration MAE is $0.286836$ for in-support values ($N=56{,}350$) and $0.348697$ out of support ($N=85{,}900$). For $C_{2017}$, seen values yield $1.01\times10^{-6}$ MAE ($N=142{,}170$), whereas unseen values yield $0.013750$ ($N=80$).

\section{Conclusion}

We studied temporal prediction under corrupted physical factors and severe target imbalance. Across four model families, Direct prediction was more robust than formula-anchored correction but remained limited in recovering rare high-magnitude changes despite modifications to the objective, representation, spatial context, and architecture.

PhyRestore instead restores corrupted physical factors before reconstructing temporal change through RUSLE. Relative to the strongest Direct baselines, it reduced TAIL95 error by 28.9\% and 76.1\% under isolated $R_{2022}$ and $C_{2017}$ corruption, respectively. Its advantage weakened under simultaneous corruption and outside the training support, demonstrating that its effectiveness depends critically on factor recoverability.

\bibliography{iclr2027_conference}

@techreport{renard1997rusle,
  author      = {Renard, Kenneth G. and Foster, George R. and Weesies, Glenn A. and McCool, Donald K. and Yoder, Douglas C.},
  title       = {Predicting Soil Erosion by Water: A Guide to Conservation Planning with the Revised Universal Soil Loss Equation ({RUSLE})},
  institution = {U.S. Department of Agriculture},
  type        = {Agriculture Handbook},
  number      = {703},
  address     = {Washington, DC, USA},
  year        = {1997},
  url         = {https://ntrl.ntis.gov/NTRL/dashboard/searchResults/titleDetail/PB97153704.xhtml}
}

@article{benavidez2018review,
  author  = {Benavidez, Rubianca and Jackson, Bethanna and Maxwell, Deborah and Norton, Kevin},
  title   = {A Review of the (Revised) Universal Soil Loss Equation (({R}){USLE}): With a View to Increasing Its Global Applicability and Improving Soil Loss Estimates},
  journal = {Hydrology and Earth System Sciences},
  volume  = {22},
  pages   = {6059--6086},
  year    = {2018},
  doi     = {10.5194/hess-22-6059-2018},
  url     = {https://doi.org/10.5194/hess-22-6059-2018}
}

@article{panagos2015assessment,
  author  = {Panagos, Panos and Borrelli, Pasquale and Poesen, Jean and Ballabio, Cristiano and Lugato, Emanuele and Meusburger, Katrin and Montanarella, Luca and Alewell, Christine},
  title   = {The New Assessment of Soil Loss by Water Erosion in Europe},
  journal = {Environmental Science \& Policy},
  volume  = {54},
  pages   = {438--447},
  year    = {2015},
  doi     = {10.1016/j.envsci.2015.08.012},
  url     = {https://doi.org/10.1016/j.envsci.2015.08.012}
}

@article{borrelli2017global,
  author  = {Borrelli, Pasquale and Robinson, David A. and Fleischer, Larissa R. and Lugato, Emanuele and Ballabio, Cristiano and Alewell, Christine and Meusburger, Katrin and Modugno, Sirio and Sch{\"u}tt, Brigitta and Ferro, Vito and Bagarello, Vincenzo and Van Oost, Kristof and Montanarella, Luca and Panagos, Panos},
  title   = {An Assessment of the Global Impact of 21st Century Land Use Change on Soil Erosion},
  journal = {Nature Communications},
  volume  = {8},
  number  = {1},
  pages   = {2013},
  year    = {2017},
  doi     = {10.1038/s41467-017-02142-7},
  url     = {https://doi.org/10.1038/s41467-017-02142-7}
}

@article{li2023soil,
  author  = {Li, Pingheng and Tariq, Aqil and Li, Qingting and Ghaffar, Bushra and Farhan, Muhammad and Jamil, Ahsan and Soufan, Walid and El Sabagh, Ayman and Freeshah, Mohamed},
  title   = {Soil Erosion Assessment by {RUSLE} Model Using Remote Sensing and {GIS} in an Arid Zone},
  journal = {International Journal of Digital Earth},
  volume  = {16},
  number  = {1},
  pages   = {3105--3124},
  year    = {2023},
  doi     = {10.1080/17538947.2023.2243916},
  url     = {https://doi.org/10.1080/17538947.2023.2243916}
}

@article{ge2023soil,
  author  = {Ge, Yuankai and Zhao, Longlong and Chen, Jinsong and Li, Xiaoli and Li, Hongzhong and Wang, Zhengxin and Ren, Yanni},
  title   = {Study on Soil Erosion Driving Forces by Using ({R}){USLE} Framework and Machine Learning: A Case Study in Southwest China},
  journal = {Land},
  volume  = {12},
  number  = {3},
  pages   = {639},
  year    = {2023},
  doi     = {10.3390/land12030639},
  url     = {https://doi.org/10.3390/land12030639}
}

@article{samarinas2024soil,
  author  = {Samarinas, Nikiforos and Tsakiridis, Nikolaos L. and Kalopesa, Eleni and Zalidis, George C.},
  title   = {Soil Loss Estimation by Water Erosion in Agricultural Areas Introducing Artificial Intelligence Geospatial Layers into the {RUSLE} Model},
  journal = {Land},
  volume  = {13},
  number  = {2},
  pages   = {174},
  year    = {2024},
  doi     = {10.3390/land13020174},
  url     = {https://doi.org/10.3390/land13020174}
}

@article{karpatne2017theory,
  author  = {Karpatne, Anuj and Atluri, Gowtham and Faghmous, James H. and Steinbach, Michael and Banerjee, Arindam and Ganguly, Auroop and Shekhar, Shashi and Samatova, Nagiza and Kumar, Vipin},
  title   = {Theory-Guided Data Science: A New Paradigm for Scientific Discovery from Data},
  journal = {IEEE Transactions on Knowledge and Data Engineering},
  volume  = {29},
  number  = {10},
  pages   = {2318--2331},
  year    = {2017},
  doi     = {10.1109/TKDE.2017.2720168},
  url     = {https://doi.org/10.1109/TKDE.2017.2720168}
}

@article{raissi2019physics,
  author  = {Raissi, Maziar and Perdikaris, Paris and Karniadakis, George Em},
  title   = {Physics-Informed Neural Networks: A Deep Learning Framework for Solving Forward and Inverse Problems Involving Nonlinear Partial Differential Equations},
  journal = {Journal of Computational Physics},
  volume  = {378},
  pages   = {686--707},
  year    = {2019},
  doi     = {10.1016/j.jcp.2018.10.045},
  url     = {https://doi.org/10.1016/j.jcp.2018.10.045}
}

@article{karniadakis2021physics,
  author  = {Karniadakis, George Em and Kevrekidis, Ioannis G. and Lu, Lu and Perdikaris, Paris and Wang, Sifan and Yang, Liu},
  title   = {Physics-Informed Machine Learning},
  journal = {Nature Reviews Physics},
  volume  = {3},
  pages   = {422--440},
  year    = {2021},
  doi     = {10.1038/s42254-021-00314-5},
  url     = {https://doi.org/10.1038/s42254-021-00314-5}
}

@article{willard2022integrating,
  author  = {Willard, Jared and Jia, Xiaowei and Xu, Shaoming and Steinbach, Michael and Kumar, Vipin},
  title   = {Integrating Scientific Knowledge with Machine Learning for Engineering and Environmental Systems},
  journal = {ACM Computing Surveys},
  volume  = {55},
  number  = {4},
  pages   = {66:1--66:37},
  articleno = {66},
  year    = {2022},
  doi     = {10.1145/3514228},
  url     = {https://doi.org/10.1145/3514228}
}

@inproceedings{vanDerMaaten2013corrupted,
  author    = {{van der Maaten}, Laurens and Chen, Minmin and Tyree, Stephen and Weinberger, Kilian},
  title     = {Learning with Marginalized Corrupted Features},
  booktitle = {Proceedings of the 30th International Conference on Machine Learning},
  series    = {Proceedings of Machine Learning Research},
  volume    = {28},
  number    = {1},
  pages     = {410--418},
  year      = {2013},
  publisher = {PMLR},
  url       = {https://proceedings.mlr.press/v28/vandermaaten13.html}
}

@inproceedings{yoon2018gain,
  author    = {Yoon, Jinsung and Jordon, James and {van der Schaar}, Mihaela},
  title     = {{GAIN}: Missing Data Imputation Using Generative Adversarial Nets},
  booktitle = {Proceedings of the 35th International Conference on Machine Learning},
  series    = {Proceedings of Machine Learning Research},
  volume    = {80},
  pages     = {5689--5698},
  year      = {2018},
  publisher = {PMLR},
  url       = {https://proceedings.mlr.press/v80/yoon18a.html}
}

@article{ribeiro2020imbalanced,
  author  = {Ribeiro, Rita P. and Moniz, Nuno},
  title   = {Imbalanced Regression and Extreme Value Prediction},
  journal = {Machine Learning},
  volume  = {109},
  pages   = {1803--1835},
  year    = {2020},
  doi     = {10.1007/s10994-020-05900-9},
  url     = {https://doi.org/10.1007/s10994-020-05900-9}
}

@inproceedings{yang2021delving,
  author    = {Yang, Yuzhe and Zha, Kaiwen and Chen, Yingcong and Wang, Hao and Katabi, Dina},
  title     = {Delving into Deep Imbalanced Regression},
  booktitle = {Proceedings of the 38th International Conference on Machine Learning},
  series    = {Proceedings of Machine Learning Research},
  volume    = {139},
  pages     = {11842--11851},
  year      = {2021},
  publisher = {PMLR},
  url       = {https://proceedings.mlr.press/v139/yang21m.html}
}

@article{stroosnijder2005measurement,
  author  = {Stroosnijder, Leo},
  title   = {Measurement of Erosion: Is It Possible?},
  journal = {CATENA},
  volume  = {64},
  number  = {2--3},
  pages   = {162--173},
  year    = {2005},
  doi     = {10.1016/j.catena.2005.08.004},
  url     = {https://doi.org/10.1016/j.catena.2005.08.004}
}

@article{boixfayos2006measuring,
  author  = {Boix-Fayos, Carolina and Mart{\'i}nez-Mena, Mar{\'i}a and Arnau-Rosal{\'e}n, Eva and Calvo-Cases, Adolfo and Castillo, Victor and Albaladejo, Juan},
  title   = {Measuring Soil Erosion by Field Plots: Understanding the Sources of Variation},
  journal = {Earth-Science Reviews},
  volume  = {78},
  number  = {3--4},
  pages   = {267--285},
  year    = {2006},
  doi     = {10.1016/j.earscirev.2006.05.005},
  url     = {https://doi.org/10.1016/j.earscirev.2006.05.005}
}

@article{phinzi2019assessment,
  author  = {Phinzi, Kwanele and Ngetar, Njoya Silas},
  title   = {The Assessment of Water-Borne Erosion at Catchment Level Using {GIS}-Based {RUSLE} and Remote Sensing: A Review},
  journal = {International Soil and Water Conservation Research},
  volume  = {7},
  number  = {1},
  pages   = {27--46},
  year    = {2019},
  doi     = {10.1016/j.iswcr.2018.12.002},
  url     = {https://doi.org/10.1016/j.iswcr.2018.12.002}
}

@ARTICLE{Ahmet_Mediterranean, 
AUTHOR={İpek, Ahmet Faruk and Kahya, Ercan },       
TITLE={Spatiotemporal prioritization of soil erosion risk using the RUSLE model and CMIP6 projections under future climate scenarios in a Mediterranean watershed},      
JOURNAL={Frontiers in Environmental Science},    
VOLUME={14},
YEAR={2026},
DOI={10.3389/fenvs.2026.1760569} 
}

@inproceedings{gong2022ranksim,
  title     = {{RankSim}: Ranking Similarity Regularization for Deep Imbalanced Regression},
  author    = {Gong, Yu and Mori, Greg and Tung, Fred},
  booktitle = {Proceedings of the 39th International Conference on Machine Learning},
  pages     = {7634--7649},
  year      = {2022},
  volume    = {162},
  url       = {https://proceedings.mlr.press/v162/gong22a.html}
}

@inproceedings{rashiwa2026segmentation,
  title     = {Segmentation Expert-Mixture Regularization:
               An Adaptive Learning Method for Imbalanced Regression Problems},
  author    = {Rashiwa, Sarah Elyane and Branco, Paula},
  booktitle = {Proceedings of the 39th Canadian Conference on Artificial Intelligence},
  pages     = {612--623},
  year      = {2026},
  volume    = {318},
  url       = {https://proceedings.mlr.press/v318/rashiwa26a.html}
}

@inproceedings{pintea2023step,
  author    = {Pintea, Silvia L. and Lin, Yancong and
               Dijkstra, Jouke and van Gemert, Jan C.},
  title     = {A Step Towards Understanding Why Classification Helps Regression},
  booktitle = {Proceedings of the IEEE/CVF International Conference on Computer Vision},
  pages     = {19915--19924},
  year      = {2023}
}

@inproceedings{vaswani2017attention,
  author    = {Vaswani, Ashish and Shazeer, Noam and Parmar, Niki and
               Uszkoreit, Jakob and Jones, Llion and Gomez, Aidan N. and
               Kaiser, {\L}ukasz and Polosukhin, Illia},
  title     = {Attention Is All You Need},
  booktitle = {Advances in Neural Information Processing Systems},
  volume    = {30},
  pages     = {5998--6008},
  year      = {2017}
}

@article{daly2002prism,
  author  = {Daly, Christopher and Gibson, Wayne and Taylor, George and Johnson, Gary and Pasteris, Paul},
  year    = {2002},
  title   = {A knowledge-based approach to the statistical mapping of climate},
  journal = {Climate Research},
  volume  = {22},
  pages   = {99--113}
}

@misc{usda_ssurgo,
  author       = {{U.S. Department of Agriculture, Natural Resources Conservation Service}},
  title        = {Soil Survey Geographic (SSURGO) Database},
  year         = {2023},
  howpublished = {\url{https://www.nrcs.usda.gov/resources/data-and-reports/soil-survey-geographic-database-ssurgo}},
  note         = {Accessed 2025-12-14}
}

@misc{usgs_3dep,
  author       = {{U.S. Geological Survey}},
  title        = {{USGS 1/3 Arc Second n39w108 20210312}},
  year         = {2021},
  howpublished = {U.S. Geological Survey, The National Map},
  note         = {Accessed 2025-12-20}
}

@INPROCEEDINGS{sentinelData,
  author={Karra, Krishna and Kontgis, Caitlin and Statman-Weil, Zoe and Mazzariello, Joseph C. and Mathis, Mark and Brumby, Steven P.},
  booktitle={2021 IEEE International Geoscience and Remote Sensing Symposium IGARSS}, 
  title={Global land use / land cover with Sentinel 2 and deep learning}, 
  year={2021},
  volume={},
  number={},
  pages={4704-4707}
}
\bibliographystyle{iclr2027_conference}


\end{document}